\documentclass[journal,twoside,web]{ieeecolor}
\usepackage{generic}
\usepackage{cite}
\usepackage{amsmath,amssymb,amsfonts}
\usepackage{graphicx}
\usepackage{booktabs}
\usepackage{array}
\usepackage{multirow}
\usepackage{makecell}
\usepackage{subcaption}
\usepackage{float}
\usepackage{hyperref}
\hypersetup{hidelinks=true}
\usepackage{textcomp}
\usepackage[utf8]{inputenc}
\usepackage{xcolor}

\def\BibTeX{{\rm B\kern-.05em{\sc i\kern-.025em b}\kern-.08em
    T\kern-.1667em\lower.7ex\hbox{E}\kern-.125emX}}

\begin{document}

\title{Early Prediction of Pathological Complete Response to
Neoadjuvant Chemotherapy Using Temporal Deep Learning on DWI}

\author{Pablo Garc\'ia Marcos,
        Md.~Tarequl Islam,
        Paula Puerta Gonz\'alez,
        Guillermo Lorenzo,
        Hector Gomez,
        Covadonga del Camino,
        Angel Rio-Alvarez,
        and V\'ictor M. Gonz\'alez%
\thanks{This work was supported in part by the Council of Gij\'on through
the University Institute of Industrial Technology of Asturias (IUTA) under
Grants SV-25-GIJON-1-14, SV-25-GIJON-1-02, SV-24-GIJON-1-05,
SV-24-GIJON-1-18, SV-24-GIJON-1-16, SV-23-GIJON-1-09, SV-22-GIJON-1-19,
and SV-21-GIJON-1-19; by Principado de Asturias under Grant
SV-PA-21-AYUD/2021/50994; and by MICIU/AEI/10.13039/501100011033 and
ESF+ under Grant RYC2022-036010-I (G.~Lorenzo).}
\thanks{P.~Garc\'ia Marcos, V.~M.~Gonz\'alez, and A.~Rio-Alvarez are with
the Computer Sciences Department and the Biomedical Engineering Center (BME),
University of Oviedo, 33003 Asturias, Spain
(e-mail: garciamarpablo@uniovi.es; vmsuarez@uniovi.es; rioangel@uniovi.es).}
\thanks{Md.~T.~Islam is with Rabindra University, Sirajganj-6770, Bangladesh
(e-mail: tareq.cse@rub.ac.bd).}
\thanks{P.~Puerta Gonz\'alez is with the Electrical Engineering Department
and the Biomedical Engineering Center (BME), University of Oviedo,
33003 Asturias, Spain (e-mail: puertapaula@uniovi.es).}
\thanks{G.~Lorenzo is with the Group of Numerical Methods in Engineering,
Department of Mathematics, University of A Coru\~na, 15071 A Coru\~na,
Spain, and also with the Oden Institute for Computational Engineering and
Sciences, The University of Texas at Austin, Austin, TX 78712 USA
(e-mail: guillermo.lorenzo@udc.es).}
\thanks{H.~Gomez is with the School of Mechanical Engineering, the Weldon
School of Biomedical Engineering, and the Purdue Center for Cancer Research,
Purdue University, West Lafayette, IN 47907 USA
(e-mail: hectorgomez@purdue.edu).}
\thanks{C.~del Camino is with the Radiodiagnostic Service, Asturias Central
University Hospital (HUCA), 33011 Oviedo, Spain
(e-mail: caminocovadonga@uniovi.es).}
\thanks{Corresponding author: Angel Rio-Alvarez (rioangel@uniovi.es).}}

\maketitle

\begin{abstract}
Early identification of non-responders to neoadjuvant chemotherapy (NACT)
is crucial for timely treatment adaptation in breast cancer. However, many
existing predictive models rely on multiparametric magnetic resonance imaging
(MRI), late treatment time points, or extensive clinical data, which limits
their applicability. This study proposes a deep learning framework for early
prediction of pathological complete response (pCR) using only
diffusion-weighted MRI (DW-MRI) acquired at baseline and after the first
NACT cycle. This framework feeds cropped tumor-centered patches to an
EfficientNet-based temporal model that directly learns tumor shape and local
tissue characteristics without explicit radiomic feature engineering. The
model, trained with 10-fold cross-validation, achieved an area under the
receiver operating characteristic curve (AUC) of 0.90 for pCR prediction
after one cycle, providing actionable information after a single treatment
cycle while avoiding gadolinium administration and reducing dependence on
heterogeneous clinical data. By focusing on the baseline-to-first-cycle
window instead of later stages, the approach supports earlier escalation or
de-escalation of NACT, and its exclusive reliance on DW-MRI facilitates
protocol standardization, multi-centre deployment and privacy-preserving data
sharing. These results demonstrate that DW-MRI-based deep learning on
tumor-centered patches constitutes a minimally invasive, clinically deployable
strategy for early pCR prediction, with direct implications for personalized
treatment adaptation in neoadjuvant breast cancer therapy.
\end{abstract}


\section{Introduction}
\label{sec:intro}

\IEEEPARstart{B}{reast} cancer remains a leading cause of morbidity and
mortality among women worldwide. Treatment has seen substantial evolution
over the past decades due to the increasing use of tailored therapies. In
this context, neoadjuvant chemotherapy (NACT) is especially relevant for
aggressive breast cancer subtypes, such as triple-negative and HER2-positive
disease. The use of NACT can reduce tumor size, facilitate surgical
interventions, provide information on treatment sensitivity, and treat
micrometastases early~\cite{Korde2021_ASCO_NACT_guideline,
gianni_neoadjuvant_nodate, von_minckwitz_neoadjuvant_2012}. One of NACT's
main objectives is to achieve pathological complete response (pCR),
characterized by the absence of residual disease in the breast and lymph
nodes. Achieving pCR is strongly associated with improved patient survival,
especially in HER2-positive and triple-negative
tumors~\cite{cortazar_pathological_2014, von_minckwitz_pcr_definition_2012,
valenza_optimizing_2024}. However, despite the therapeutic advances, a
significant percentage of patients do not achieve pCR, exposing themselves
to the toxicity of NACT without fully improving their
condition~\cite{esserman_pathologic_2012, bianchini_triple-negative_2016}.

This context highlights the importance of early evaluation of NACT response.
Predictive models capable of estimating treatment response and the likelihood
of achieving pCR at early treatment stages could allow for effective
treatment tailoring, improving clinical outcomes and reducing unnecessary
toxicity~\cite{Tudorica2016_EarlyPrediction_DCE,
Liu2021_EarlyPrediction_SciRep, esserman_pathologic_2012}. This personalized
approach could improve outcomes and patient quality of life. Nevertheless,
the focus of a substantial set of scientific literature has been the
evaluation of NACT response during advanced treatment stages, which can be
too late to include meaningful changes in the
therapy~\cite{Korde2021_ASCO_NACT_guideline}. Despite the success of these
studies predicting the pathological response, their clinical utility remains
limited. Conversely, predictive models designed for early treatment stages
are of particular interest for improving clinical
workflows~\cite{hylton_neoadjuvant_2016}.

Magnetic resonance imaging (MRI) has become an important reference technique
for the assessment and monitoring of breast cancer because it provides
detailed information about tumor volume and tissue
characteristics~\cite{Korde2021_ASCO_NACT_guideline}. MRI protocols for
NACT of breast cancer predominantly rely on dynamic contrast-enhanced MRI
(DCE-MRI). This technique employs a contrast agent to capture enhancement
patterns over time, which tend to be altered by the abnormal vasculature of
tumor tissue. Other MRI techniques are used as complementary tools in modern
medical workflows, such as diffusion-weighted MRI (DW-MRI), which is based
on water diffusion~\cite{hylton_neoadjuvant_2016,
partridge_diffusion-weighted_2018}. MRI techniques are commonly used to
measure the residual disease after NACT and plan the subsequent surgical
intervention. Nevertheless, research has shown a gap between the residual
disease found in MRI and the achievement of pCR, with significant variability
in sensitivity and specificity~\cite{marinovich_meta-analysis_2013}.
Furthermore, the MRI-based predictive accuracy of pCR varies widely between
the different molecular subtypes~\cite{lobbes_role_2013}.

Despite the lesion conspicuity of tumorous tissue offered by DCE-MRI, its
dependence on contrast agents based on gadolinium creates clinical and
logistical issues, including higher cost, longer scan times, and additional
personnel and material resources required for administration and
monitoring~\cite{Mann2024_NonContrastBreastMRI}. Additionally, the use of
contrast poses further safety concerns for the patients, as they may suffer
adverse reactions, particularly in patients with severe renal
impairment~\cite{blomqvist_gadolinium_2022, cunningham_safety_2024,
starekova_update_2024}. Finally, it should be noted that the medical imaging
obtained can differ widely between centers, hindering standardization and the
generalization of models~\cite{mann_breast_2019}. Together, these limitations
make multicenter adoption of DCE-MRI-based predictive models more
challenging. Besides the use of standalone DCE-MRI, many studies have used
multimodal inputs, combining DCE-MRI with the use of DW-MRI, clinical
information and biomarkers. Despite the potential improvement in performance,
this approach increases the dependency on heterogeneous data that may vary
between medical centers, hindering reproducibility and its real use in
standard medical workflows~\cite{Tudorica2016_EarlyPrediction_DCE,
Liu2021_EarlyPrediction_SciRep}. Recent research highlights the issues of
multimodal pipelines with the integration of different clinical inputs,
requiring strict standardization and validation to avoid the degradation of
results outside the original domain~\cite{mohamed_multiparametric_2024}.
These issues motivate simpler predictive strategies that depend less on
heterogeneous clinical data and are easier to reproduce and standardize. The
logistical burden associated with contrast-enhanced protocols also motivates
exploring imaging strategies that do not require contrast administration.

DW-MRI enables non-invasive characterization of breast tissue by measuring
water-molecule motion, often restricted in tumor-affected tissue. Apparent
diffusion coefficient (ADC) maps are also routinely created from DW-MRI,
offering insight into tissue properties across different b-values used to
measure diffusion. For breast cancer, evidence suggests that DW-MRI and
ADC's information can be used both for diagnosis and monitoring response to
NACT. In particular, longitudinal changes in ADC have been associated with
pathologic response during
NACT~\cite{Partridge2018_ACRIN6698, Partridge2023_DWI_update}. Hence,
DW-MRI is seen as a promising alternative to DCE-MRI, as avoiding the use
of contrast agents enables simpler and safer medical protocols. Recent
literature has shown the potential of DW-MRI both for patient assessment and
response measurement, making it a more appealing option, together with
standardization improvements~\cite{Pereira2019_DWI_breast_currentStatus,
Mann2024_NonContrastBreastMRI}. Thus, the use of standalone DW-MRI (or its
derived ADC maps) could offer a simpler and safer alternative for the
creation of prediction models. Reduced dependence on heterogeneous inputs and
streamlined acquisition may ease adoption and implementation across multiple
medical centers~\cite{Partridge2018_ACRIN6698, Partridge2023_DWI_update,
Mann2024_NonContrastBreastMRI}.

The objective of this study is to develop a predictive model to obtain early
predictions of pCR at surgery using DW-MRI acquired at baseline and after
completion of the first NACT cycle. To this end, only DW-MRI data and its
derived ADC maps will be used, thereby maximizing clinical applicability and
ease of adoption~\cite{Tudorica2016_EarlyPrediction_DCE,
Liu2021_EarlyPrediction_SciRep}. For the prediction task, we propose to use
Deep Learning (DL) to create a family of computer vision models capable of
learning pCR response from longitudinal imaging changes measured on patches
of tumor-affected tissue. This approach is grounded in the proven utility of
DW-MRI/ADC to monitor NACT response and the increasing interest in
contrast-free MRI techniques~\cite{Partridge2018_ACRIN6698,
Partridge2023_DWI_update, Mann2024_NonContrastBreastMRI}. Furthermore,
recent research has explored the use of automatic evaluation and patients'
response, highlighting the need to investigate diffusion-based pipelines for
clinical applications~\cite{Gilad2022_PDDWI_arxiv,
DWIonly2024_AutomatedResponse}. In summary, the main contribution of this
work is the development of a standalone diffusion MRI-based predictive
pipeline for early NACT response in breast cancer patients, designed to
exploit longitudinal tumor changes through temporal DL modeling.

The rest of this paper is structured as follows.
Section~\ref{sec:background} reviews state-of-the-art computer vision
techniques in medical imaging. Section~\ref{sec:matdata} describes the
dataset used in this work and our DL models, including their architecture,
implementation details and the metrics used to define their performance. In
Section~\ref{sec:results}, we present and discuss the results of our study.
Finally, Section~\ref{sec:conclusion} provides final conclusions and future
avenues of research.

\section{Related Work}
\label{sec:background}

Patient-response prediction to NACT using MRI has been explored with a wide
range of techniques. Early research saw the use of radiomics together with
machine learning (ML) algorithms. In these studies, quantitative descriptors
were extracted from various sequences, such as DCE-MRI or DW-MRI. These
descriptors were later used to train supervised ML models to predict
pCR~\cite{Tudorica2016_EarlyPrediction_DCE, danna_comparative_2025}. This
paradigm demonstrated that metrics derived from DCE and diffusion can capture
information relevant to treatment response. The use of ML algorithms showed
clinically relevant results, especially when combined with clinical
data~\cite{Tudorica2016_EarlyPrediction_DCE, wang_mri-based_2025}.
Nevertheless, despite their contributions as ``proof of concept'', these
radiomics-ML pipelines present practical limitations. First, this approach
often relies on manual tumor segmentation, greatly increasing processing time
and introducing variability due to the human factor, hindering
generalization~\cite{khan_deep_2022}. Additionally, radiomic features exhibit
high variability depending on scanner vendor and acquisition protocol, further
complicating multicenter generalization without reliable harmonization
strategies~\cite{danna_comparative_2025}. These studies also tend to use
multiparametric models and heterogeneous clinical information, increasing the
complexity of the workflow and requiring data that may not be available in all
medical institutions~\cite{Liu2021_EarlyPrediction_SciRep,
wang_mri-based_2025}. These factors have motivated interest in simpler
approaches that improve reproducibility and multicenter generalization.

Early DL approaches for NACT response prediction from MRI commonly used
DCE-MRI as the primary input. In this framework, various convolutional neural
network (CNN) architectures were trained using pretreatment or post-NACT
images to predict pCR, improving in many cases the results achieved by
radiomics with ML~\cite{khan_deep_2022, Liu2021_EarlyPrediction_SciRep}.
Recent models display promising AUC and sensitivity, with potential use as a
support tool for oncologists' decision-making. Nevertheless, these models
tend to focus on late timepoints (post-NACT or near treatment completion),
which limits their usefulness for treatment
adaptation~\cite{khan_deep_2022, Liu2021_EarlyPrediction_SciRep}. Also,
despite the methodological innovation, these approaches present significant
limitations. Beyond the clinical and logistical constraints of
gadolinium-based contrast agents, DL models are sensitive to domain shifts
caused by image acquisition and DCE phase selection. These limitations
generate increasing interest in contrast-free alternative techniques (such
as DW-MRI) and the integration of biomarkers~\cite{Mann2024_NonContrastBreastMRI,
khan_deep_2022, Partridge2023_DWI_update}.

The evolution of DL models based on a single data type created a new wave of
approaches integrating multiple sources of information to improve pCR
prediction. In this new multimodal paradigm, models combine multiple sequence
types with clinical data and, in some cases, histological data and biomarkers.
The use of multiple data types allows models to leverage complementary
information to improve the AUC and
sensitivity~\cite{joo_multimodal_2021, Liu2021_EarlyPrediction_SciRep}.
Nevertheless, this performance gains come with increased pipeline complexity
and data requirements. In real clinical environments, differences in data
acquisition between centers can reduce model performance. Furthermore,
heterogeneity in the coding and availability of histological data could
further hinder model deployment. These limitations are consistent with recent
reviews showcasing that, despite the improved performance, this approach
increases the need for standardization, and is particularly vulnerable to
domain shifts~\cite{Mann2024_NonContrastBreastMRI, Partridge2023_DWI_update,
khan_deep_2022}.

Various recent studies have given DW-MRI a primary role in the prediction of
NACT response, shifting emphasis toward diffusion-based biomarkers derived
from this technique and ADC maps. Early works in this direction included the
use of radiomics derived from ADC maps and the development of DL models based
on DW-MRI combined with other MRI sequences and clinical data. In this
context, the evidence obtained from the multicenter trial ACRIN~6698
demonstrates the potential of ADC maps for NACT response
predictions~\cite{Partridge2018_ACRIN6698}. From this foundation, multiple
studies have explored DW-MRI, generally integrating its information with
other input sources. One example is PD-DWI, a ML model that leverages
clinical data and DW-MRI to predict NACT response, showcasing the potential
of these inputs with DL models crafted to capture longitudinal
changes~\cite{Gilad2022_PDDWI_arxiv}. In parallel, recent reviews focused on
DW-MRI for breast cancer summarize accumulated evidence on its applicability
for diagnosis, characterization and response evaluation, together with the
challenges and limitations of its clinical
implementation~\cite{Partridge2023_DWI_update}. Nevertheless, despite the
potential of DW-MRI displayed by these studies, most DL models do not rely on
DW-MRI as the sole input. Instead, recent DL models usually integrate
DCE-MRI, clinical or histological data to improve performance, introducing
dependence on heterogeneous data. Thus, there is a lack of strategies that
leverage standalone DW-MRI, aiming for an equilibrium between predictive
performance and ease of deployment.

In recent years, some studies have tested the use of standalone DW-MRI (and
diffusion-derived biomarkers such as ADC) during early treatment stages.
Given the aforementioned potential of ADC for NACT response predictions,
recent reviews support the utility of diffusion-related techniques for
treatment response monitoring~\cite{Partridge2018_ACRIN6698,
Partridge2023_DWI_update}. Nevertheless, despite this recent interest, most
of the works based only on diffusion present recurrent limitations, such as
the competitive disadvantage against models with additional information. In
this context, multiple studies have explored the viability of early prediction
with standalone MRI~\cite{DWIonly2024_AutomatedResponse,
richard_diffusion-weighted_2013, chu_diffusion-weighted_2018}. Regardless,
literature reviews suggest a methodological gap in the generation of early
predictions through DL using just diffusion images gathered from multiple
centers, as in our work.

In summary, scientific literature has progressively advanced from pipelines
based on radiomics and ML to more complex DL models, in many cases
multimodal. Nevertheless, this progress has been tied to relevant
trade-offs: such as the dependence on contrast and the complexity of
integrating multiple sequences. In parallel, although evidence supports the
prediction capabilities of diffusion for response monitoring, many approaches
use it as a complementary input in multimodal implementations or in
predictions near treatment end, limiting its impact and
applicability~\cite{Partridge2018_ACRIN6698, Partridge2023_DWI_update,
Mann2024_NonContrastBreastMRI}. In this context, our work positions itself
to fill a clear gap: use exclusively DWI at baseline and after one NACT
cycle, learning from tumor-centered patches. This approach aims to offer
early predictive tools with lower logistical load and easier
standardization~\cite{Tudorica2016_EarlyPrediction_DCE,
Liu2021_EarlyPrediction_SciRep, Gilad2022_PDDWI_arxiv,
DWIonly2024_AutomatedResponse}.

\section{Materials and Methods}
\label{sec:matdata}

\subsection{Data Description}

\begin{figure}[!t]
\centerline{\includegraphics[width=\columnwidth]{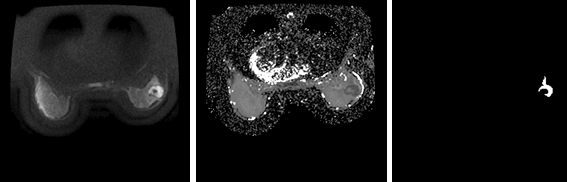}}
\caption{Slice of a study of the dataset in DW (left), ADC (center) and
mask (right) formats.}
\label{fig:dataset}
\end{figure}

The dataset ACRIN~6698 was used for model training and evaluation. This
publicly available dataset was obtained from The Cancer Imaging
Archive~\cite{noauthor_acrin_nodate}. ACRIN~6698 was released as part of a
multicenter clinical trial focused on NACT for breast cancer, seeking to
evaluate the potential of ADC maps to predict NACT
response~\cite{newitt_testretest_2019, partridge_diffusion-weighted_2018}.
The dataset includes imaging studies of $n=385$ patients, each with up to
four MRI examinations acquired at key timepoints: before treatment (T0),
after 3~weeks of treatment with paclitaxel (T1), between paclitaxel and
anthracycline (T2) and after 4~anthracycline cycles (T3). In each MRI
session, multiple sequences were obtained, including DCE-MRI, DW-MRI, ADC
maps, and expert manual tumor segmentations for DW-MRI and ADC. These
segmentations are represented as binary masks, where the tumor is the
positive class and the negative class corresponds to all other tissues.
Fig.~\ref{fig:dataset} shows a slice of DW-MRI, ADC map, and tumor
segmentation for a patient. Details on MRI acquisition have been provided in
previous works~\cite{newitt_testretest_2019, partridge_diffusion-weighted_2018}.

\subsection{Pre-Processing}

\begin{figure}[!t]
\centerline{\includegraphics[width=0.85\columnwidth]{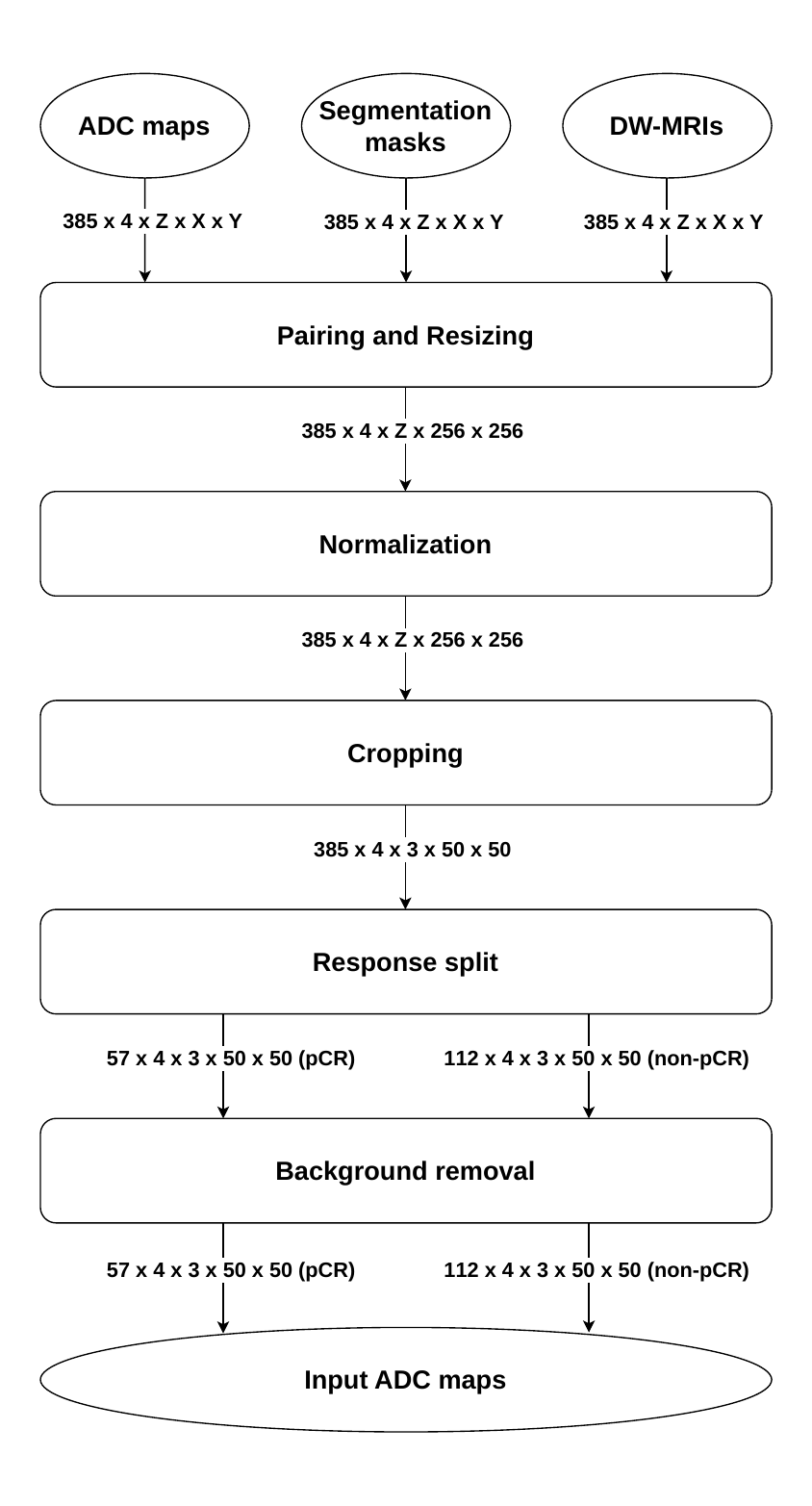}}
\caption{Pre-processing pipeline. Values of the arrows represent
$\text{Patients}\times\text{Time points}\times\text{Slices}\times
\text{Width}\times\text{Height}$.}
\label{fig:pre}
\end{figure}

In this work, data was curated through a pre-processing pipeline. First,
each ADC and DW slice was paired with its corresponding segmentation mask and
resized to $256\times256$. Pairing was performed using the metadata ID numbers
from each patient as reference, ensuring a match in labels across tumor masks,
ADC maps, and DW-MRI data. Then, volumes were reconstructed from the
individual slices and normalized to the interval $[0,1]$ through a contrast
stretch, clipping at the 99.8th percentile (top~0.2\%) and rescaling to
$[0,1]$; see~\cite{kamnitsas_efficient_2017, zhao_deep_2018}. Lower and
higher limit values for normalization were defined globally for the dataset.
During this volume reconstruction step, background (black) masks were
generated for those DW-MRI/ADC slices without a matching mask in the dataset.
These slices corresponded to regions without visible tumor according to the
annotation protocol. The volumes were then cropped to shape $3\times50\times50$
by selecting the 3D patch that maximized the number of positive voxels in the
mask. The volumes were cropped to 3~consecutive slices to allow using them as
input for DL models pretrained on RGB images. The $50\times50$ slice size was
empirically selected, maximizing the tumor tissue in the image without
excessively reducing the size of the latter. Subsequently, patients were split
by response (112~non-pCR and 57~pCR), following the dataset's metadata.
During this split, to ensure data reliability and consistency, all patients
without all diffusion studies were removed from the pipeline. Finally, all
non-tumorous tissue in the ADC images was masked out (set to zero). The
non-tumorous tissue was defined as all pixels labeled as background in the
segmentation masks, excluding a 5-pixel margin around the tumor borders. This
margin of 5~pixels was determined through empirical evaluation between 3,~5
and~7 pixels. Fig.~\ref{fig:pre} shows the pipeline steps and the
dimensional changes in the dataset produced by each step.

\subsection{Methods}

This section describes the deep learning algorithms used for feature
extraction and NACT response prediction. In all cases, the input images are
ADC maps.

\subsubsection{Deep Learning Architectures}

\begin{figure}[!t]
\centerline{\includegraphics[width=0.85\columnwidth]{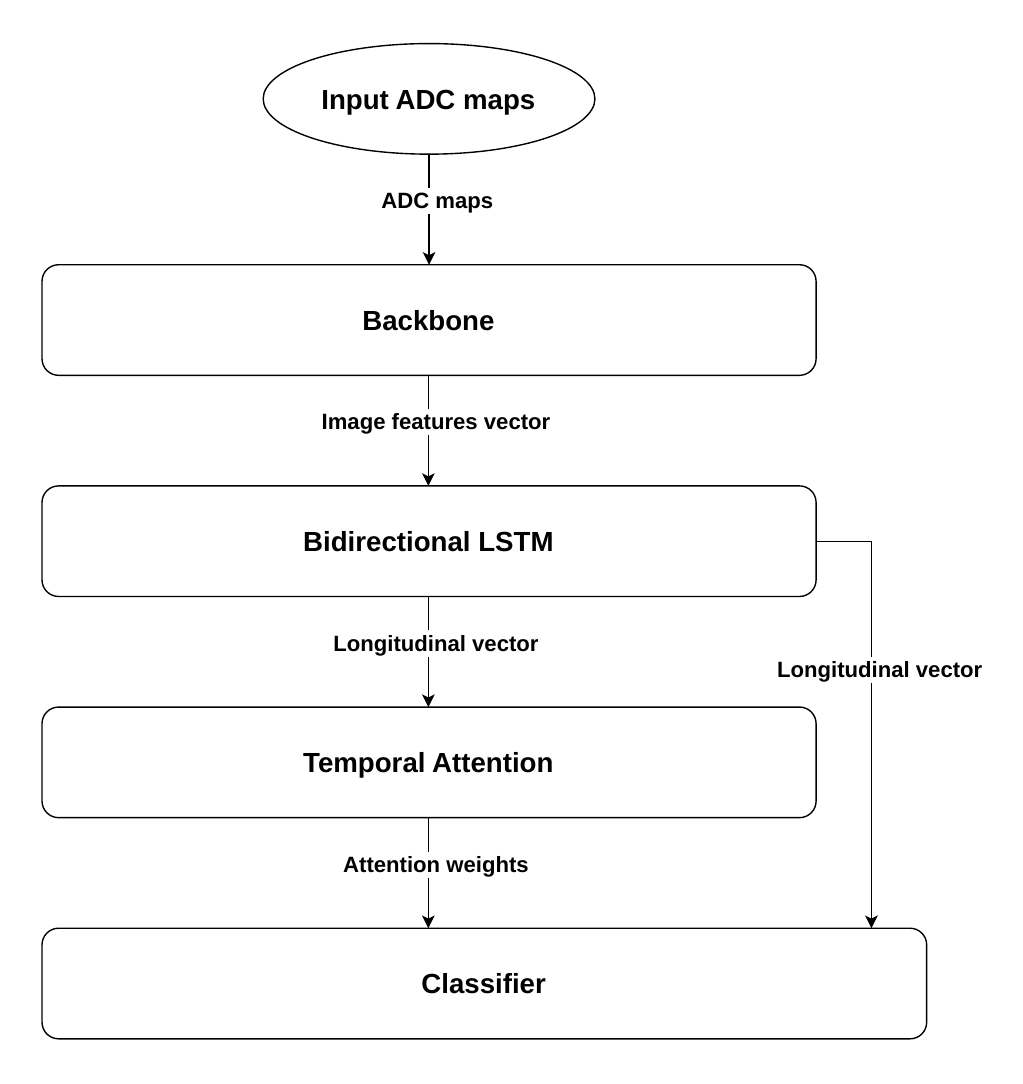}}
\caption{NACT response prediction model architecture.}
\label{fig:model}
\end{figure}

Feature extraction is performed using a CNN as the backbone (feature
extractor). In particular, we use the EfficientNet-B0 architecture with
pre-trained weights. Extracted features were then used by a recurrent neural
network (RNN) to generate longitudinal information of the changes between the
different timepoints. Specifically, a bidirectional long short-term memory
(LSTM) network was used, a type of RNN designed to model temporal sequences
and explore them in both directions. The information generated by the LSTM is
then enriched with a temporal attention module, tasked with adjusting each
timepoint's weight. This temporal attention module employs two linear layers
and an activation to produce a score for each timepoint followed by a softmax
normalization across timepoints. This results in a set of attention weights
that quantify the relative importance of each timepoint, allowing the
classification model to focus on the most informative stage of the treatment.
Finally, a classifier employs the information generated by the LSTM network,
the temporal attention, and the quality indices associated with the images to
perform a binary classification. Quality indices are categorical and numerical
metadata describing the assessed quality of each patient's data. This
prediction is done using two linear layers, using ReLU activation and a
dropout function of~40\%. This neural network architecture is hereafter
referred to as the final model. Fig.~\ref{fig:model} provides a visual
representation of the proposed final model.

\subsubsection{Training, Validation and Test Data}

Model training was run for up to 100~epochs, with early stopping patience set
to 50~epochs. The Adam optimizer was used with a learning rate of $10^{-4}$,
reduced by a factor of~0.5 after 10~epochs without improvement in validation
accuracy. The loss function used was cross-entropy. We used 169~patients
(112~non-pCR and 57~pCR) as input, leveraging the T0 and T1
$3\times50\times50$ images for each. For model-development ablations,
patients were split using $k$-fold cross-validation with $k=5$, with
approximately 80\% of patients for training and the remainder for validation.
For the final training and reporting of the model, $k$-fold cross-validation
with $k=10$ was used instead, employing 90\% of patients for training. Class
imbalance was addressed by oversampling the minority class through the
duplication of pCR patient samples to match the number of non-pCR samples.
Data augmentation was applied on-the-fly to training data through random
horizontal and vertical flipping (applied with probabilities of 0.5 and~0.2,
respectively), rotations ($\pm10^{\circ}$), translations of up to~5\% and
scaling of up to $\pm$5\%. Mild intensity perturbations were also introduced
by adjusting brightness and contrast within range~[0.9,~1.1], in both cases.

\subsubsection{Performance Metrics}

Model performance was evaluated by comparing model predictions against
ground-truth annotations. Classification performance was quantified using
accuracy, F1-score, and area under the receiver operating characteristic
curve (AUC-ROC). AUC-ROC measures the model's discriminative ability across
all decision thresholds. We used the following definitions:

\begin{equation}
\text{Accuracy} = \frac{\text{TP} + \text{TN}}{\text{TP} + \text{TN} + \text{FP} + \text{FN}}
\label{eq:accuracy}
\end{equation}

\begin{equation}
\text{Recall} = \frac{\text{TP}}{\text{TP} + \text{FN}}
\label{eq:recall}
\end{equation}

\begin{equation}
\text{Precision} = \frac{\text{TP}}{\text{TP} + \text{FP}}
\label{eq:precision}
\end{equation}

\begin{equation}
\text{F1-score} = 2 \times \frac{\text{Precision} \times \text{Recall}}{\text{Precision} + \text{Recall}}
\label{eq:f1}
\end{equation}

Here, TP, TN, FP and FN indicate, respectively, true positive, true
negative, false positive and false negative for the binary prediction of
NACT response, with pCR being defined as the positive class.

\section{Results}
\label{sec:results}

\subsection{Ablation Study}
\label{subsec:ablation}

\subsubsection{Temporal Modeling and Backbone}

During our experimental process, we first compared a non-recurrent temporal
model with a recurrent LSTM-based model. We performed this comparison using
multiple pre-trained backbone combinations. This comparison was used to assess
whether modeling temporal dynamics is beneficial when only limited
longitudinal information (T0--T1) is available, compared to a non-recurrent
approach based on features from T0, T1, and their difference. In the
non-recurrent temporal setting, features extracted independently at T0 and T1
are concatenated and passed through fully connected layers, under the initial
assumption that a lightweight aggregation could suffice with just two temporal
snapshots. Recurrent modeling systematically achieved better results than the
lightweight aggregation.

Table~\ref{tab:ablation_backbones_nohist} summarizes the backbone and
temporal modeling ablation for the T0--T1 setting using 5-fold
cross-validation on $50\times50$ tumor patches. All configurations use
shape-based radiomics only, without quality indices.

\begin{table}[!t]
\caption{Backbone Ablation on the T0--T1 Setting Using 5-Fold
Cross-Validation on $50\times50$ Tumor Patches. Mean Validation Accuracy
and F1-Score Across Folds for Different Backbones and Temporal Modules
(No Quality Indices).}
\label{tab:ablation_backbones_nohist}
\centering
\begin{tabular}{lccc}
\toprule
Backbone & Temporal & Acc (\%) & F1 \\
\midrule
EfficientNet-B0  & LSTM          & 76.0 & 0.77 \\
EfficientNet-B0  & Non-recurrent & 75.0 & 0.76 \\
EffNet-B3 (RA)   & LSTM          & 74.0 & 0.74 \\
EffNet-B3 (RA)   & Non-recurrent & 71.0 & 0.72 \\
EffNetV2-B3      & LSTM          & 72.0 & 0.72 \\
EffNetV2-B3      & Non-recurrent & 72.0 & 0.72 \\
DenseNet-121     & LSTM          & 70.0 & 0.71 \\
DenseNet-121     & Non-recurrent & 69.0 & 0.70 \\
ViT (Base)       & LSTM          & 68.0 & 0.69 \\
ViT (Base)       & Non-recurrent & 67.0 & 0.68 \\
\bottomrule
\end{tabular}
\end{table}

Across all backbones, the LSTM-based temporal modeling either matches or
outperforms the non-recurrent temporal aggregation, indicating that even with
only two time points, the model benefits from explicitly learning temporal
dynamics. The best overall performance is obtained with the EfficientNet-B0
backbone combined with the LSTM temporal module, reaching 76\% accuracy and
0.77~F1, while the corresponding simple temporal version remains slightly
below (75\% accuracy, 0.76~F1).

\subsubsection{Temporal Module Hyperparameters}

After selecting EfficientNet-B0 with an LSTM-based temporal module as the
preferred backbone configuration, we conducted a second ablation to determine
the optimal LSTM hyperparameters. Specifically, we explored different hidden
state dimensionalities and dropout rates while keeping all other components
fixed. The goal of this stage was to identify a configuration that maximizes
the validation F1-score while preserving stability across folds.

Table~\ref{tab:ablation_lstm_hidden_dropout} reports the mean validation
accuracy and F1-score for each LSTM configuration. The results show a
consistent trend in which increasing either hidden size or dropout slightly
improves performance, with the best scores obtained for 64~hidden units and
a dropout rate of~0.40. This configuration achieves the highest mean F1-score
while avoiding signs of overfitting, and is therefore adopted as the temporal
module in the model used for the subsequent experiments.

\begin{table}[!t]
\caption{Ablation on LSTM Hidden Size and Dropout Rate for the
EfficientNet-B0 Backbone in the T0--T1 Setting Using 5-Fold
Cross-Validation on $50\times50$ Tumor Patches.}
\label{tab:ablation_lstm_hidden_dropout}
\centering
\begin{tabular}{lcc}
\toprule
LSTM configuration & Acc (\%) & F1 \\
\midrule
32 units, dropout 0.30 & 74.0 & 0.74 \\
32 units, dropout 0.40 & 75.0 & 0.76 \\
64 units, dropout 0.30 & 75.0 & 0.76 \\
64 units, dropout 0.40 & 76.0 & 0.77 \\
\bottomrule
\end{tabular}
\end{table}

\subsubsection{Quality Indices}

Besides the image features extracted by the convolutional backbone from the
$50\times50$ tumor-centered patches (including the 5-pixel margin around the
lesion), we considered a set of image quality indices that quantify different
aspects of the reliability of the underlying MRI images and segmentations
(e.g.\ signal-to-noise properties, motion-related artifacts, and temporal
consistency of the tumor region). These indices are computed at lesion level
and concatenated to the feature representation produced by the backbone and
temporal module, with the aim of modulating the prediction according to the
confidence in the acquired data.

To preliminarily assess their impact, we performed an ablation using the
selected EfficientNet-B0 backbone and LSTM temporal module on the T0--T1
setting with 5-fold cross-validation.
Table~\ref{tab:ablation_qc_5fold} reports the mean validation accuracy and
F1-score when using only the imaging features from the cropped patches versus
the same features augmented with the quality indices.

\begin{table}[!t]
\caption{Effect of Adding Image Quality Indices (QC) to the Imaging
Features Extracted from $50\times50$ Tumor-Centered Patches
(T0--T1, EfficientNet-B0 + LSTM), Using 5-Fold Cross-Validation.}
\label{tab:ablation_qc_5fold}
\centering
\begin{tabular}{lcc}
\toprule
Feature set & Acc (\%) & F1 \\
\midrule
Imaging features only     & 76.0 & 0.77 \\
Imaging + quality indices & 77.0 & 0.77 \\
\bottomrule
\end{tabular}
\end{table}

Including the quality indices produces an increase of approximately one
percentage point in mean accuracy, while the F1-score remains essentially
unchanged. Although this is a positive effect, the difference is small and
not conclusive at this ablation stage, which is based on a limited 5-fold
evaluation. For this reason, and given the clinical interpretability of the
quality indices, we decided not to discard them but to re-evaluate their
impact under the final optimized configuration. The following discussion of
the results delves into the impact of quality indices in the final model.

\subsection{Experimental Results and Analysis}

\begin{figure}[!t]
\centerline{\includegraphics[width=\columnwidth]{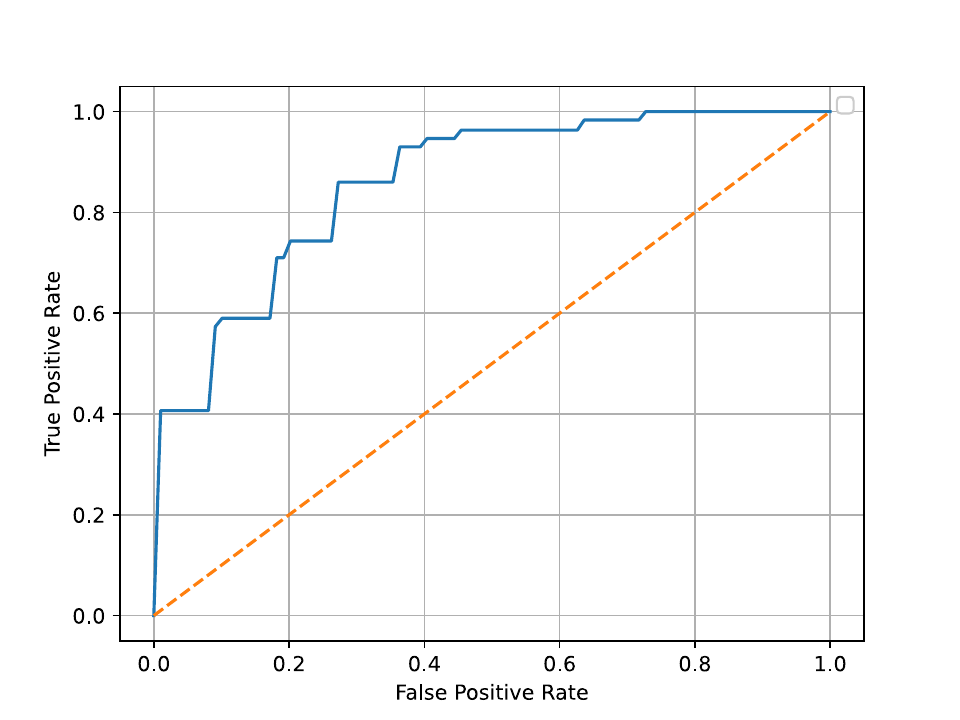}}
\caption{ROC curve showing the average performance across the 10
cross-validation folds.}
\label{fig:AUC_ROC}
\end{figure}

Table~\ref{tab:final_per_fold} reports the per-fold validation performance
of the final model used for early-response prediction with only the first two
treatment time points (T0 and~T1). Individual fold analysis shows high
inter-fold variability, likely due to the limited size of the validation
subset in each fold, where the classification outcome of each patient can
strongly influence the performance metrics. Fig.~\ref{fig:AUC_ROC} reports
the AUC-ROC of the model, computed as the average performance across the
10~cross-validation folds.

\begin{table}[!t]
\caption{Per-Fold Validation Accuracy (Acc), F1-Score and AUC for the Final
T0--T1 Model with Image Quality Indices (10-Fold Cross-Validation,
$50\times50$ Tumor-Centered Patches).}
\label{tab:final_per_fold}
\centering
\begin{tabular}{cccc}
\toprule
Fold & Acc (\%) & F1 & AUC \\
\midrule
1  & 88.24 & 0.88 & 0.92 \\
2  & 94.12 & 0.94 & 0.97 \\
3  & 88.24 & 0.88 & 0.86 \\
4  & 70.59 & 0.71 & 0.76 \\
5  & 88.24 & 0.88 & 0.91 \\
6  & 82.35 & 0.83 & 0.94 \\
7  & 87.50 & 0.88 & 0.88 \\
8  & 87.50 & 0.88 & 0.96 \\
9  & 75.00 & 0.76 & 0.91 \\
10 & 81.25 & 0.82 & 0.84 \\
\midrule
Mean $\pm$ SD & $84.30\pm7.08$ & $0.85\pm0.07$ & $0.90\pm0.06$ \\
\bottomrule
\end{tabular}
\end{table}

The use of EfficientNet-B0 seemed to provide a good balance between
representational capacity and regularization for this relatively small
dataset, avoiding the overfitting tendencies that can appear with deeper or
more complex backbones such as EffNet-B3 or DenseNet. Moreover, the
bi-directional LSTM allowed exploiting subtle intensity and shape changes
between T0 and T1, capturing patterns that a simple concatenation of features
may not model adequately.

In contrast, the ViT-based backbone consistently yielded the lowest scores
among the tested architectures, both with non-recurrent and LSTM temporal
modules. This behavior is consistent with previous observations that vision
transformers typically require substantially larger datasets and stronger data
augmentation to reach their full potential, whereas in small-sample
medical-imaging scenarios convolutional backbones tend to be more
effective~\cite{takahashi_comparison_2024}.

To assess the specific contribution of the quality indices, we additionally
trained the final model without them, keeping the same 10-fold partitions. In
this setting, the model without quality indices achieved a mean validation
accuracy of 82.24\% ($\text{SD}\approx8.33$) and a mean F1-score of~0.83
($\text{SD}\approx0.08$), whereas the version including quality indices
reached 84.3\% ($\text{SD}\approx8.18$) accuracy and 0.85
($\text{SD}\approx0.07$) F1-score. Although the absolute differences are
moderate, the configuration with quality indices consistently matches or
improves the performance across the majority of folds in the 10-fold
cross-validation and yields a higher mean accuracy and F1-score.

Despite the difficulties in drawing direct and fair comparisons due to the
use of different inputs and techniques, some studies from the literature are
close enough to be comparable. Table~\ref{tab:comparison} summarizes the
closest research articles, together with a schematic overview of their
approaches and results. Reference~\cite{partridge_diffusion-weighted_2018}
delved into the use of diffusion for NACT response prediction, achieving an
AUC of~0.61, while~\cite{choi_early_2020} reported an AUC of~0.75 using ADC
from up to three weeks as the sole input. On the other hand,
references~\cite{mani_early_nodate, liu_radiomics_2019} used earlier time
points (pre-treatment or first week), achieving an AUC of~0.96 and~0.86,
respectively, but they utilized also DCE data and clinical variables.
Similarly, multiparametric approaches combining DCE and DWI have reported
AUCs of~0.86~\cite{tahmassebi_impact_2019, liu_radiomics_2019}. Our proposed
predictive model achieves a competitive performance in early prediction of
pCR (AUC~$=0.90$), in a cross-validation framework, exceeding ADC-based
approaches and slightly improving upon some multiparametric methods, although
remaining below~\cite{mani_early_nodate}. This suggests that a prediction
focusing on the tumor, using only DW-MRI, may be comparable to more complex
models, while offering substantial simplicity for multicenter deployment.

\begin{table*}[!t]
\caption{Comparison of Previous Studies and the Proposed Approach for Early
Prediction of pCR to NACT in Breast Cancer. Only MRI-Related Metrics Are
Reported for Papers Employing Multiple Approaches.}
\label{tab:comparison}
\centering
\setlength{\tabcolsep}{6pt}
\begin{tabular}{p{2.5cm}p{6.5cm}p{3.5cm}c}
\toprule
\textbf{Study} & \textbf{Input Data} & \textbf{Prediction Timepoint} & \textbf{AUC} \\
\midrule
\cite{partridge_diffusion-weighted_2018} & DWI & Mid-treatment (12 weeks) & 0.61 \\[2pt]
\cite{choi_early_2020} & DWI & Early treatment (3 weeks) & 0.75 \\[2pt]
\cite{tahmassebi_impact_2019} & DCE + DWI + T2 (weighted MRI sequence) & After 2 NACT cycles & 0.86 \\[2pt]
\cite{liu_radiomics_2019} & DCE + DWI + T2 (weighted MRI sequence) + Clinical variables & Pre-treatment & 0.86 \\[2pt]
\cite{mani_early_nodate} & DCE + DWI + Clinical variables (only 20 samples) & After 1 NACT cycle & 0.96 \\[2pt]
\midrule
Current & DWI & 1 NACT cycle & 0.90 \\
\bottomrule
\end{tabular}
\end{table*}

\section{Conclusions and Future Work}
\label{sec:conclusion}

The results of this study highlight the feasibility of early NACT-response
prediction using only information before treatment onset and during the first
cycle of treatment. The use of a temporal model based on EfficientNet-B0 +
bidirectional LSTM with patches centered in the tumor achieved an accuracy of
$84.30\pm7.08$, a F1-score of $0.85\pm0.07$ and an AUC of $0.90\pm0.06$
across 10~folds of cross-validation. These findings confirm the reliability
of the approach for early treatment response prediction while maintaining
methodological simplicity. The achieved AUC surpasses several comparable
approaches and remains competitive with respect to multimodal models (see
Table~\ref{tab:comparison}). Individual fold analysis reveals moderate
performance variability, with a notably weaker fold. This suggests that
performance may be affected by partition composition. Nevertheless, most
folds present strong performances (AUC typically above~$\approx0.88$),
supporting the robustness of the approach in the validation scheme. Moreover,
the ablation study suggests that the incorporation of quality indices
improves, although slightly, the metrics achieved by the predictor. The
average gain (approximately one percentage point) suggests the usefulness of
these descriptors as complementary information to refine predictions. Finally,
the multicenter nature of the dataset suggests a strong potential for
generalization to new clinical centers without substantial performance
degradation. Overall, these results highlight the usefulness of
diffusion-based tumor patches for early prediction of NACT response and
support further evaluation in multicenter clinical settings.

Future work may further explore extensions of the present framework. Although
using data up to the end of the first NACT cycle could support oncologists'
decision-making, predictions could be generated earlier using input from just
T0 (before the first cycle), generating early response estimations prior to
systemic treatment exposure. These T0 predictions could be further improved
using clinical variables, such as patient age or cancer subtype. Furthermore,
the use of additional imaging sequences could be incorporated to aid
performance. Nevertheless, any performance gains from using additional data
must be carefully weighed against the previously discussed disadvantages of
incorporating new heterogeneous inputs. Finally, external validation on
independent centers could assess generalization and robustness in real-world
implementations.

While these findings are encouraging, some limitations should be considered.
The final cohort (169~patients) is relatively small, which may limit the
statistical robustness of the results, as reflected in the observed inter-fold
variability. In addition, the reliance on expert-annotated masks for tumor
delineation reduces clinical scalability and highlights the need for automated
segmentation approaches.

\section*{Acknowledgment}

This research was supported in part by the Council of Gij\'on through the
University Institute of Industrial Technology of Asturias (IUTA) under Grants
SV-25-GIJON-1-14, SV-25-GIJON-1-02, SV-24-GIJON-1-05, SV-24-GIJON-1-18,
SV-24-GIJON-1-16, SV-23-GIJON-1-09, SV-22-GIJON-1-19, and
SV-21-GIJON-1-19; by Principado de Asturias under Grant
SV-PA-21-AYUD/2021/50994; and by MICIU/AEI/10.13039/501100011033 and ESF+
under Grant RYC2022-036010-I (G.~Lorenzo).

\section*{Ethics Statement}

All procedures in this research were conducted in compliance with applicable
laws and institutional guidelines. This study exclusively used the ACRIN~6698
dataset, a publicly available, fully de-identified imaging dataset of breast
cancer patients obtained from The Cancer Imaging Archive (TCIA) under
accession number \texttt{tcia.kk02-6d95}~\cite{noauthor_acrin_nodate},
available at \url{https://doi.org/10.7937/tcia.kk02-6d95}. In accordance
with TCIA data use policies, Institutional Review Board (IRB) approval and
informed consent were not required for this secondary analysis of
de-identified, publicly available data. The authors confirm that no ethical
complications were identified during the course of the study.

\section*{Declaration of Competing Interest}

The authors declare that they have no known competing financial interests or
personal relationships that could have appeared to influence the work reported
in this paper.

\bibliographystyle{IEEEtran}
\bibliography{bibliography}

\begin{IEEEbiography}[{\includegraphics[width=1in,height=1.25in,clip,keepaspectratio]{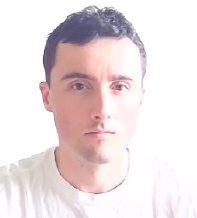}}]{Pablo Garc\'ia Marcos}
received his B.Sc. degree in Software Engineering in 2019 and his M.Sc. degree in Web Engineering in 2025, both from the University of Oviedo, Asturias, Spain. He worked as a Full-Stack Engineer at Arvo Consultores y Tecnolog\'ia S.L. (2018--2021) and at the European Organization for Nuclear Research (CERN) from 2021 to 2023. Since 2023, he has been a Lecturer and Researcher at the University of Oviedo, where he is pursuing a Ph.D. in biomedical applications of computer vision. His research interests include neuron culture characterization, breast tumor detection and segmentation, and medical image analysis.
\end{IEEEbiography}

\begin{IEEEbiography}[{\includegraphics[width=1in,height=1.25in,clip,keepaspectratio]{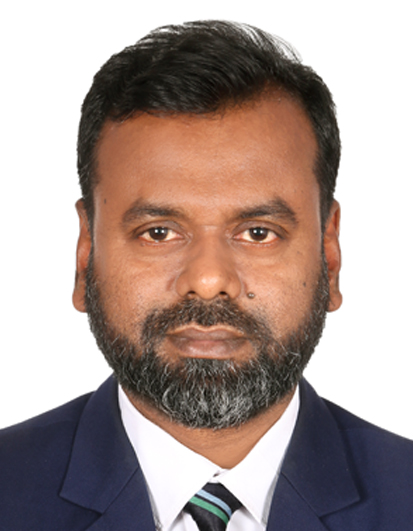}}]{Md. Tarequl Islam}
received the B.Sc. degree in Computer Science and Engineering from the University of Rajshahi, Bangladesh, and the M.Sc. (Engg.) degree from Mawlana Bhasani Science and Technology University, Bangladesh, where he is currently pursuing the Ph.D. degree. He completed a research stay at the Biomedical Engineering Center, Universidad de Oviedo, Spain, in 2025. He has held positions at Grameenphone, LM Ericsson Bangladesh, and Khwaja Yunus Ali University, and is currently an Assistant Professor at Rabindra University, Bangladesh. He has authored more than 27 publications indexed by IEEE, Elsevier, Springer, and Taylor \& Francis. His research interests include artificial intelligence, computer vision, deep learning, and medical image analysis.
\end{IEEEbiography}

\begin{IEEEbiography}[{\includegraphics[width=1in,height=1.25in,clip,keepaspectratio]{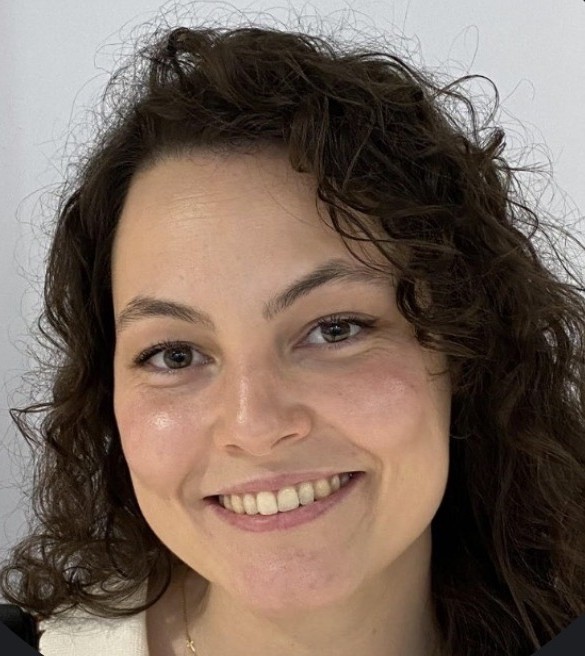}}]{Paula Puerta Gonz\'alez}
received the B.Sc. degree in Software Engineering from the University of Oviedo, Spain, in 2023, and the M.Sc. degree in Artificial Intelligence from the University of A Coru\~na, Spain, in 2025, receiving honors for her thesis on breast cancer segmentation using deep learning and MRI. She is currently pursuing the Ph.D. degree in Computer Science at the University of Oviedo. She has worked as a Research Assistant in AI at the University of Oviedo and as a Researcher in Computer Vision developing DL models for breast cancer MRI analysis. Since 2026, she is a Data Specialist in the Health Data Space at FINBA, Oviedo, Spain. Her research interests include computer vision, medical image analysis, health data interoperability, and trustworthy AI for healthcare.
\end{IEEEbiography}

\begin{IEEEbiography}[{\includegraphics[width=1in,height=1.25in,clip,keepaspectratio]{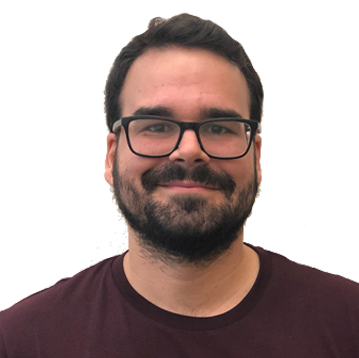}}]{Guillermo Lorenzo}
received the Ph.D. degree in Civil Engineering from the University of A Coru\~na, Spain, in 2018. He held postdoctoral positions at the University of Pavia, The University of Texas at Austin (Peter O'Donnell Jr. Fellow and Marie Sk\l{}odowska-Curie Fellow), and the Health Research Institute of Santiago de Compostela (Fundaci\'on ``La Caixa'' Fellow). He is currently a Ram\'on y Cajal Fellow at the University of A Coru\~na and a Research Affiliate at the Oden Institute, UT Austin. His awards include an ERC Starting Grant (2026), the Juan Carlos Sim\'o Prize from SEMNI (2025), and a Leonardo Scholarship from the BBVA Foundation (2025). His research interests include computational oncology, digital twins in medicine, and mechanistic learning.
\end{IEEEbiography}

\begin{IEEEbiography}[{\includegraphics[width=1in,height=1.25in,clip,keepaspectratio]{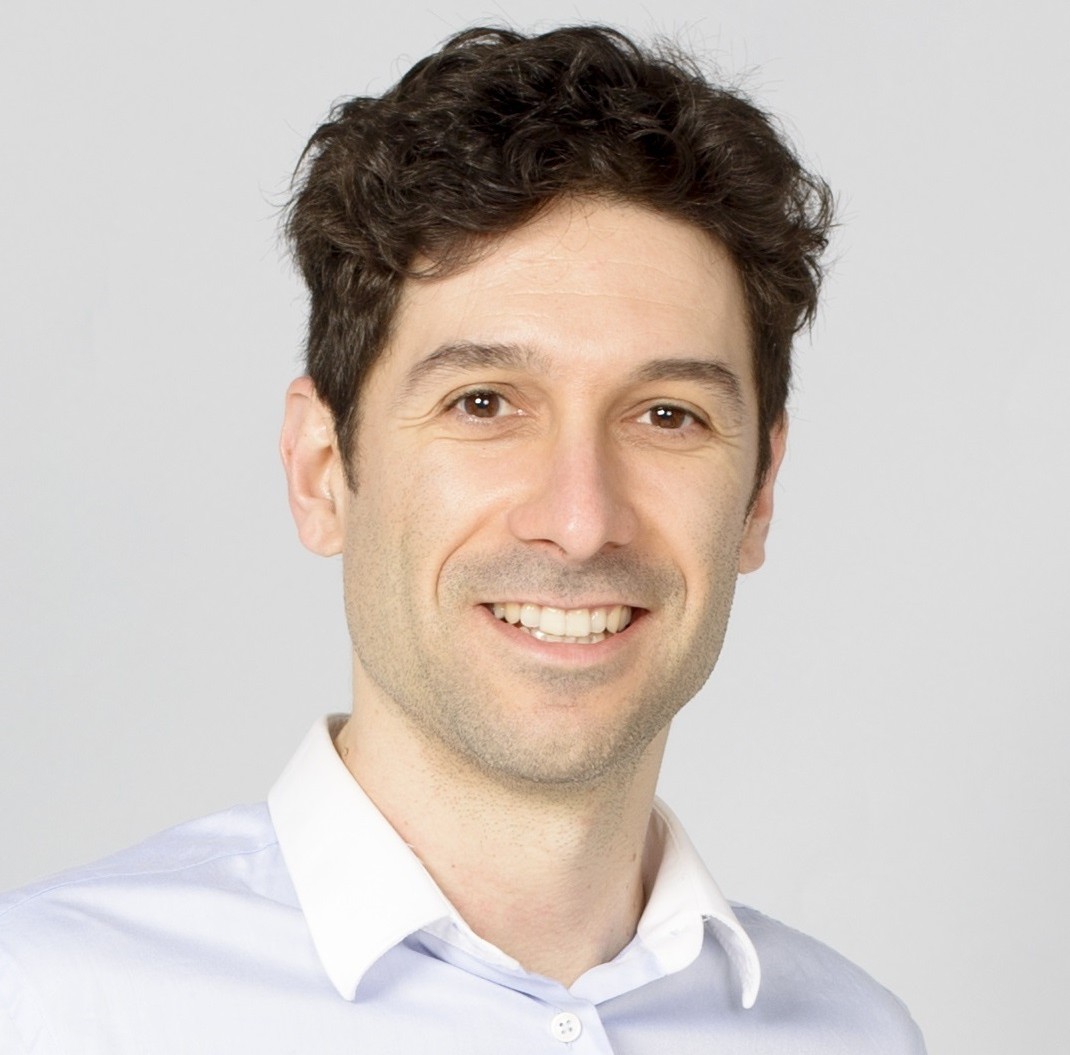}}]{Hector Gomez}
received the Ph.D. degree in Civil Engineering from the University of A Coru\~na, Spain, in 2006. He is the Morris Goldman Chair in Engineering and a Professor with the School of Mechanical Engineering, Purdue University, West Lafayette, IN, USA, with courtesy appointments in the Weldon School of Biomedical Engineering and the Purdue Institute for Cancer Research. He has published more than 120 journal articles and 200 conference contributions. His honors include the MIT Innovators Under 35 Award, the Princess of Girona Scientific Research Award, and the Fellow Award from the U.S. Association for Computational Mechanics. His research interests include computational mechanics, isogeometric modeling, and simulation at the interface of engineering and medicine.
\end{IEEEbiography}

\begin{IEEEbiography}[{\includegraphics[width=1in,height=1.25in,clip,keepaspectratio]{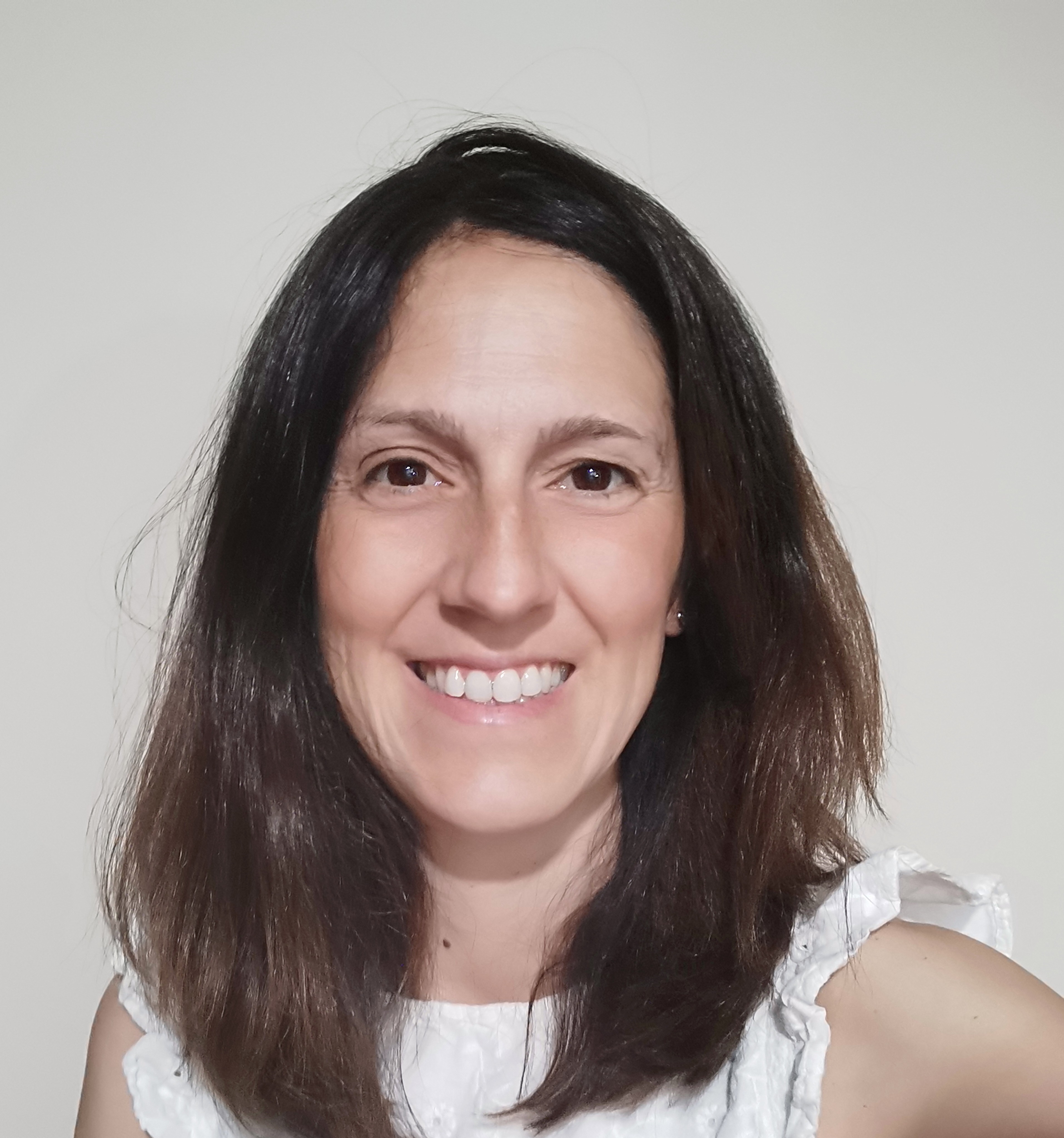}}]{Covadonga del Camino}
received her medical degree and Ph.D. in Medicine from the Universidad de Oviedo, Spain, in 2002 and 2017, respectively, and completed her residency in Diagnostic Radiology at the Hospital Universitario Central de Asturias (HUCA) in 2007. She is currently a Consultant Radiologist in the Breast Imaging Unit at HUCA, specializing in screening mammography, breast ultrasound, breast MRI, and image-guided interventional procedures. She has also served as a part-time professor at the Universidad de Oviedo. Her research interests focus on advanced multimodal breast oncology imaging. She is a member of SERAM, SEDIM, and EUSOBI.
\end{IEEEbiography}

\begin{IEEEbiography}[{\includegraphics[width=1in,height=1.25in,clip,keepaspectratio]{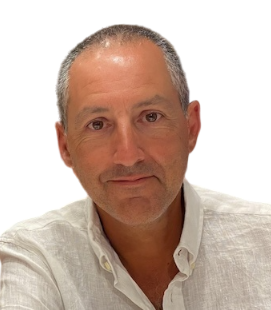}}]{V\'ictor M. Gonz\'alez}
received the B.Sc., M.Sc., and Ph.D. degrees in Computer Science from the University of Oviedo, Spain, in 1993, 1997, and 2002, respectively. He is currently an Associate Professor in the Department of Electrical Engineering and Head of the Biomedical Engineering Center at the University of Oviedo, where he leads the MIDAS Research Group. He has held visiting appointments at the University of Sk\"ovde, Rutgers University, and Purdue University. He served as Deputy-Dean for Internationalization of the Polytechnic School of Engineering of Gij\'on for 10 years and as Vice-President of the Spanish Society of Artificial Intelligence in Biomedicine. His research interests include AI, deep learning, biomedical image analysis, neuroengineering, and precision medicine.
\end{IEEEbiography}

\begin{IEEEbiography}[{\includegraphics[width=1in,height=1.25in,clip,keepaspectratio]{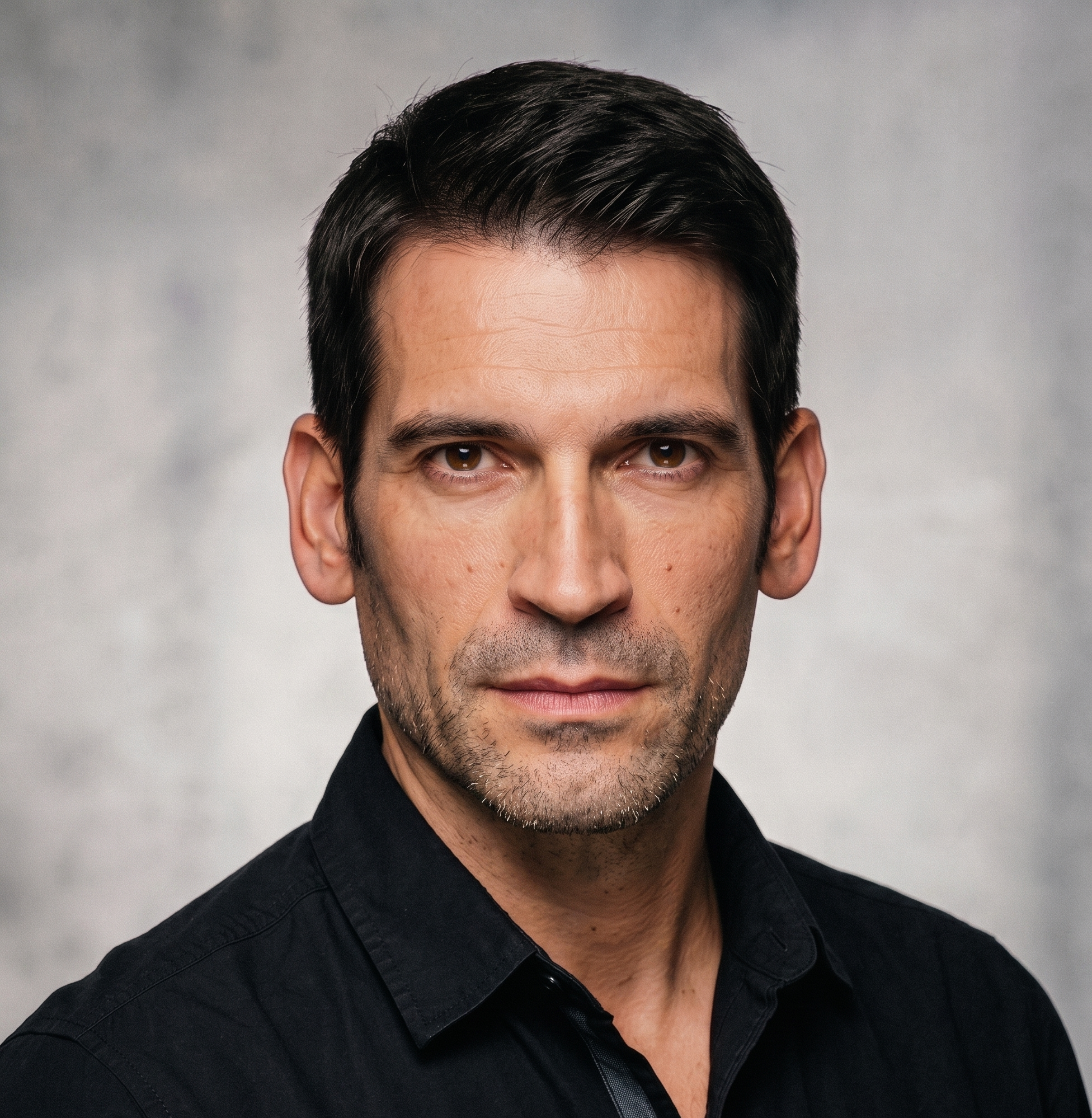}}]{Angel Rio-Alvarez}
received the B.Sc. degree in Management Informatics Engineering, the M.Sc. degree in Web Engineering, and the Ph.D. degree in Computer Science from the University of Oviedo, Oviedo, Spain, in 2012, 2015, and 2020 respectively. He is currently a Lecturer and Researcher with the Department of Computer Science, University of Oviedo, where he coordinates the Computer Vision Area of the Biomedical Engineering Research Group. His research interests include medical image analysis, computer vision for neuronal identification and characterization in cell cultures, breast cancer imaging, brain tumor detection, myocardial fibrosis assessment, deep learning, and intelligent healthcare systems. He is a member of IABiomed, SEIB, BME, and IUTA, and serves as a reviewer for international journals and conferences in artificial intelligence and biomedical engineering.
\end{IEEEbiography}

\end{document}